%% file: arxiv.tex
\documentclass[letterpaper, 10 pt, conference]{ieeeconf}  

\usepackage{color-edits}
\usepackage[dvipsnames]{xcolor}
\definecolor{cmured1}{HTML}{C41230}
\definecolor{cmured2}{HTML}{941120}

\title{\LARGE \bf
Learning Versatile Humanoid Manipulation with Touch Dreaming
}

\author{Yaru Niu$^{1,3}$, Zhenlong Fang$^{1}$, Binghong Chen$^{1}$,
Shuai Zhou$^{1}$, Revanth Krishna Senthilkumaran$^{1,3}$, \\
 Hao Zhang$^{1,2}$, Bingqing Chen$^{3}$, Chen Qiu$^{3}$, H. Eric Tseng$^{2}$, Jonathan Francis$^{1,3}$, and Ding Zhao$^{1}$
\vspace{1mm}\\
$^{1}$Carnegie Mellon University, $^{2}$UT Arlington, $^{3}$Bosch Center for AI
\vspace{1mm}\\
\href{https://humanoid-touch-dream.github.io/}{\textcolor{cmured2}{humanoid-touch-dream.github.io}}
}

\usepackage{multirow}
\usepackage{booktabs}
\usepackage{subcaption}
\usepackage{algorithm}
\usepackage{algorithmic}

\usepackage{pifont}
\usepackage{hyperref}
\usepackage[dvipsnames]{xcolor}
\usepackage{graphicx}
\usepackage{float}
\usepackage{amsmath}
\usepackage{adjustbox}
\usepackage{array}
\usepackage{amsfonts}
\usepackage{amssymb}
\usepackage{bm}
\usepackage{cite}
\usepackage[font=small]{caption}
\newcommand{\parab}[1]{\noindent\textbf{#1}}

\let\labelindent\relax
\usepackage[inline]{enumitem}

\begin{document}

\thispagestyle{empty}
\pagestyle{empty}

\twocolumn[{
\renewcommand\twocolumn[1][]{#1}
\maketitle
\vspace{-0.15in}
\begin{center}
\noindent\begin{minipage}{1.0\textwidth}
    \includegraphics[width=1.0\linewidth]{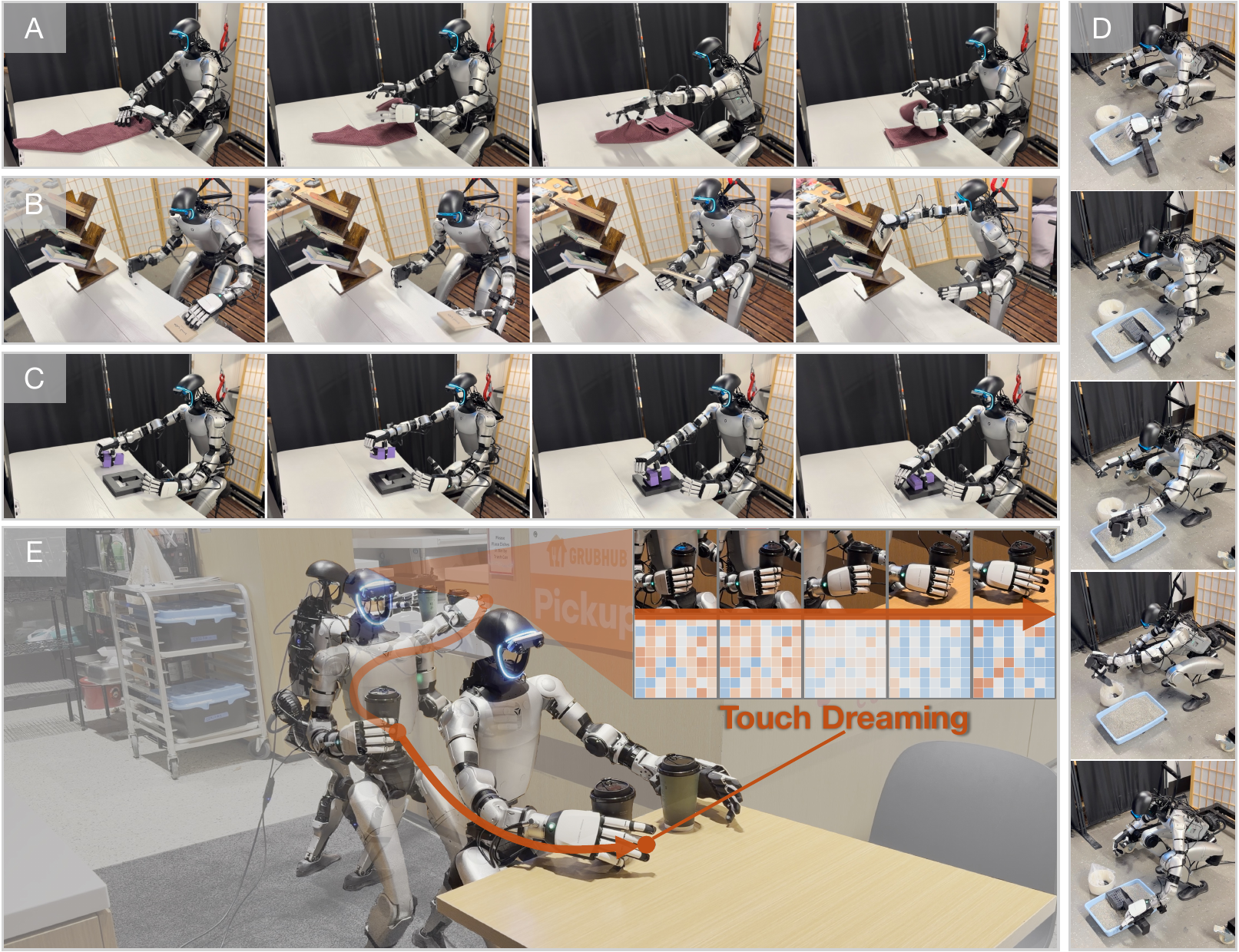}
    \end{minipage}\hspace{0.05in}
    \vspace{-0.1cm}
    \captionof{figure}{
    Our system enables \textbf{versatile}, \textbf{contact-rich}, and \textbf{dexterous} humanoid manipulation. \textit{A:} long-horizon, multi-stage manipulation of deformable objects (towel folding). \textit{B:} mixed prehensile and non-prehensile manipulation for thin-profile rigid objects with limited grasp affordance (book organization). \textit{C:} tight-tolerance insertion with a clearance of 3.5\,mm, requiring high precision and reactive adaptation (Insert-T). \textit{D:} dexterous, tool-mediated contact under low-profile constraints (cat litter scooping). \textit{E:} bimanual object fetch and loco-manipulation, requiring stable whole-body transport while keeping objects balanced (tea serving); the embedded panel visualizes \textbf{touch dreaming} by showing the dreamed tactile latent of the right middle finger as a normalized heatmap that changes with finger-object contact.
    }
    \label{fig:head}
    \vspace{-0.1cm}
\end{center}
}]

\begin{abstract}
	\input{arxiv_contents/abstract}
\end{abstract}


\section{Introduction}
\label{sec:introduction}

\input{arxiv_contents/introduction}

\section{Related Work}
\label{sec:related_work}

\input{arxiv_contents/related_work}

\section{Methodology}
\label{sec:methodology}
\input{arxiv_contents/methodology}

\section{Experiments}
\label{sec:experiments}
	\input{arxiv_contents/experiments}

\section{Conclusions}
\label{sec:conclusions}
	\input{arxiv_contents/conclusions}



\section*{Acknowledgements}
This work was partly supported by NSF SCC under Grant No.~2531320. We thank Yuxiang Yang and Peide Huang for their insightful discussions, guidance, and feedback throughout this project. We also thank Howard Han for his assistance in repairing the dexterous hands.


\bibliographystyle{IEEEtran}
\bibliography{references}

\clearpage

\appendix
\label{sec:appendix}
	\input{arxiv_contents/appendix}

\end{document}

%% file: arxiv_contents/abstract.tex
Humanoid robots promise general-purpose assistance, yet real-world humanoid loco-manipulation remains challenging because it requires \emph{whole-body stability}, \emph{end-effector dexterity}, and \emph{contact-aware interaction} under frequent contact changes. In this work, we study \textbf{dexterous, contact-rich humanoid loco-manipulation}. We first develop an RL-based \textbf{lower-body controller} that serves as the stability backbone for whole-body execution during complex manipulation. Building on this controller, we develop a VR-based \textbf{whole-body humanoid data collection system} that integrates dexterous hands and tactile sensing for contact-rich manipulation. We then propose \textbf{Humanoid Transformer with Touch Dreaming (HTD)}, a multimodal encoder--decoder Transformer that models touch as a core modality alongside multi-view vision and proprioception. HTD is trained in a single stage with behavioral cloning augmented by \textit{touch dreaming}: in addition to predicting action chunks, the policy predicts future hand-joint forces and future tactile latents, with tactile-latent targets provided by an exponential moving average target encoder without requiring a separate tactile pretraining stage. This encourages the policy to learn contact-aware representations for dexterous manipulation. Across five real-world contact-rich tasks, HTD achieves a \textbf{90.9\%} relative improvement in average success rate over the stronger baseline for each task. Ablation results further show that latent-space tactile prediction is more effective than raw tactile prediction, yielding a \textbf{30\%} relative gain in success rate. These results demonstrate that our touch-dreaming-enhanced learning system enables versatile, high-dexterity humanoid manipulation in the real world. More information and open-source materials are available at \href{https://humanoid-touch-dream.github.io/}{\textcolor{cmured2}{humanoid-touch-dream.github.io}}.

%% file: arxiv_contents/introduction.tex
Humanoid robots promise general-purpose physical assistance, fueled by rapid progress in whole-body control, teleoperation, and humanoid learning systems~\cite{he2024omnih2o, liao2025beyondmimic, wu2026perceptive, yang2025omniretarget, fu2024humanplus, luo2025sonic}. Yet real-world humanoid loco-manipulation remains fundamentally challenging because it requires tight coordination between whole-body stability and end-effector dexterity under frequent contact changes. Unlike fixed-base manipulation, humanoid manipulation is physically coupled with torso posture, base motion, and foot-ground stability, so uncertain hand-object contact can affect not only local manipulation accuracy but also whole-body execution. Consequently, accurate hand motion alone is insufficient; reliable humanoid manipulation requires stable whole-body coordination together with contact-aware interaction.

Despite these requirements, few existing systems provide a practical real-world pipeline that jointly supports stable whole-body execution, full dexterous-hand control, and tactile sensing. Although recent humanoid systems have improved motion tracking, teleoperation, and demonstration collection~\cite{ze2025twist2, li2026omniclone, zhu2026clot}, Table~\ref{tab:comparisons} highlights that few systems combine \emph{whole-body control}, \emph{full end-effector dexterity}, and \emph{touch sensing/modeling} in a single platform for dexterous, contact-rich manipulation. To address this, we build an integrated \textbf{whole-body humanoid manipulation system} that combines VR teleoperation with an RL-based lower-body controller (LBC), upper-body IK, dexterous hand retargeting, distributed tactile sensing, and touch modeling. This design provides a stable execution backbone for collecting high-quality real-world demonstrations while allowing the operator to focus on task intent and dexterous interaction.

With this tactile-enabled system in place, the next challenge is how to learn contact-aware policies from multimodal humanoid demonstrations. Purely action-supervised behavioral cloning from vision and proprioception can struggle in contact-rich manipulation, where contact is partially observed and changes abruptly over time~\cite{lee2020making}. Tactile sensing provides a natural complementary signal for capturing these contact dynamics, and prior work has demonstrated its value in visuo-tactile manipulation and predictive tactile learning~\cite{calandra2018more, heng2025vitacformer, ye2025learning, zheng2026omnivta}. Yet most existing tactile learning methods are developed for arm-hand manipulation and often rely on separate tactile pretraining, explicit world-model modules, multi-stage inference, or manually designed virtual targets tied to specific sensing and control setups~\cite{zhao2024transferable, ye2025learning, zheng2026omnivta, yuan2026vtam, chen2025implicitrdp}.
More broadly, predictive latent learning in Joint-Embedding Predictive Architectures such as I-JEPA~\cite{assran2023self} and V-JEPA~2~\cite{assran2025v} suggests that future prediction in latent space can induce semantically meaningful representations without reconstructing raw observations or training a separate generative pipeline. However, these ideas have rarely been brought into a single-stage whole-body humanoid imitation policy that must jointly handle dexterous manipulation, locomotion-related action generation, and rapidly changing contact.

Motivated by this gap, we propose \textbf{Humanoid Transformer with Touch Dreaming (HTD)}, a multimodal encoder--decoder Transformer for dexterous humanoid loco-manipulation. HTD models touch as a core modality alongside multi-view vision and proprioception, and is trained in a \emph{single stage} with behavioral cloning augmented by \textbf{touch dreaming}. In addition to predicting action chunks, HTD predicts future hand-joint forces and future tactile latents. The tactile targets are produced by an exponential moving average (EMA) target encoder, yielding stable latent supervision without requiring a separate tactile pretraining stage. Rather than using future touch prediction as a separate world model or inference-time module, HTD uses it as an auxiliary objective that regularizes the shared Transformer trunk to learn contact-aware latent dynamics while keeping deployment simple.

We evaluate the full system on five real-world contact-rich tasks: \textit{Insert-T}, \textit{Book Organization}, \textit{Towel Folding}, \textit{Cat Litter Scooping}, and \textit{Tea Serving}. These tasks span tight-tolerance insertion, hard-to-grasp rigid-object manipulation, long-horizon deformable-object handling, low-profile tool use, and bimanual loco-manipulation, together stressing precise alignment, sustained contact, whole-body coordination, and diverse interaction modes. Across these tasks, HTD achieves a \textbf{90.9\%} relative improvement in average success rate over the stronger ACT baseline for each task, while ablations show that latent tactile dreaming is more effective than raw tactile prediction. 
Together, these results suggest that combining stable whole-body execution, tactile-enabled dexterous manipulation, and predictive touch dreaming provides a practical path toward reliable humanoid manipulation under frequent contact changes.
Our contributions are threefold:
\begin{itemize}
    \item We develop a tactile-enabled whole-body humanoid manipulation system for stable, dexterous, contact-rich real-world manipulation.
    
    \item We introduce Humanoid Transformer with Touch Dreaming (HTD), a multimodal encoder--decoder Transformer for humanoid loco-manipulation that treats touch as a core modality and is trained in a single stage with touch dreaming, including future hand-joint-force prediction and EMA-supervised tactile-latent prediction for contact-aware representation learning.
    
    \item We evaluate the full framework on five real-world contact-rich humanoid manipulation tasks and show that HTD consistently improves over baselines, with latent tactile prediction outperforming raw tactile prediction.
\end{itemize}

%% file: arxiv_contents/related_work.tex
\begin{table}[htp]
\setlength{\tabcolsep}{3pt}
\centering
\caption{Comparisons to previous humanoid manipulation learning systems}
\label{tab:comparisons}
\resizebox{\columnwidth}{!}{%
\small
\setlength{\tabcolsep}{7pt}
\begin{tabular}{c c c c c}
\toprule
\multirow{2}{*}{Method} & \multirow{2}{*}{\begin{tabular}[c]{@{}c@{}} End-Effector \\ Dexterity \end{tabular}} & \multirow{2}{*}{\begin{tabular}[c]{@{}c@{}} Whole- \\ Body \end{tabular}} & \multirow{2}{*}{\begin{tabular}[c]{@{}c@{}} Touch \\ Sensing \end{tabular}} & \multirow{2}{*}{\begin{tabular}[c]{@{}c@{}} Touch \\ Modeling \end{tabular}}  \\ 
&&&& \\
\midrule
OmniH2O \cite{he2024omnih2o} & Dex-Hand Full & \ding{51} & \ding{55} & \ding{55} \\
HumanPlus \cite{fu2024humanplus} & Dex-Hand Full & \ding{51} & \ding{55} & \ding{55} \\
Mobile-TeleVision \cite{lu2025mobile} & Dex-Hand Full & \ding{51} & \ding{55} & \ding{55} \\
AMO~\cite{li2025amo} & Dex-Hand Full & \ding{51} & \ding{55} & \ding{55} \\
ViTacFormer~\cite{heng2025vitacformer} & Dex-Hand Full & \ding{55} & \ding{51} & \ding{51} \\
TWIST2~\cite{ze2025twist2} & Dex-Hand Open/Close & \ding{51} & \ding{55} & \ding{55} \\
ViTac Humanoid~\cite{kwon2025humanoid} & Dex-Hand Full & \ding{55} & \ding{51} & \ding{55} \\
SONIC~\cite{luo2025sonic} & Dex-Hand Open/Close & \ding{51} & \ding{55} & \ding{55} \\
Humanoid UMI~\cite{nai2026humanoidmanipulationinterfacehumanoid} & Gripper & \ding{51} & \ding{55} & \ding{55} \\
HumDex~\cite{heng2026humdex} & Dex-Hand Full & \ding{51} & \ding{55} & \ding{55} \\
\textbf{Ours} & Dex-Hand Full & \ding{51} & \ding{51} & \ding{51} \\
\bottomrule
\end{tabular}
}
\end{table}

\subsection{Humanoid Whole-Body Control and Teleoperation for Manipulation}

Recent progress in humanoid manipulation has been enabled by advances in whole-body control, motion tracking, and teleoperation infrastructure. A central question in humanoid whole-body control is how to represent and execute task commands across diverse behaviors, including locomotion, loco-manipulation, and upper-body manipulation. Prior works instantiate different control interfaces, such as root tracking, joint-space tracking, and body-keypoint or pose tracking, depending on the task and operator interface~\cite{he2024learning, fu2024humanplus, cheng2024open, he2024omnih2o}. One line of work improves robustness through decomposition, separating functions such as lower-body stabilization, upper-body tracking, force adaptation, or compliance modulation, as in dual-agent force-adaptive control~\cite{zhang2025falcon}, heterogeneous meta-control over multiple control modes~\cite{wei2025hmc}, adaptive compliance control~\cite{chen2025chip}, and hybrid optimization-and-learning frameworks for dexterous whole-body behaviors~\cite{li2025amo, lu2025mobile}. Related systems also combine learned whole-body control with specialized teleoperation hardware or tracking modules for more precise loco-manipulation~\cite{cheng2024expressive, ben2025homie}. Another line instead seeks unified controllers that directly coordinate locomotion and manipulation within a single whole-body tracking framework~\cite{sun2025ulcunifiedfinegrainedcontroller, ze2025twist}. Complementary teleoperation and motion-tracking systems further improve the practicality and scalability of commanding humanoids through RGB- or pose-based shadowing, immersive VR interfaces, portable mocap-free setups, and closed-loop long-horizon tracking~\cite{fu2024humanplus, he2024learning, he2024omnih2o, cheng2024open, ze2025twist2, nai2026humanoidmanipulationinterfacehumanoid, luo2025sonic, li2025clone, li2026omniclone, zhu2026clot}. Building on this line of work, our system combines an RL-based lower-body controller with a VR-based teleoperation stack using a unified reference frame, upper-body IK, and hand retargeting, enabling efficient collection of whole-body humanoid manipulation demonstrations for downstream policy learning.

\subsection{Imitation Learning for Humanoid Manipulation}


Building on these advances, humanoid manipulation systems have used real-world whole-body teleoperation to collect trajectories for imitation learning~\cite{fu2024humanplus, he2024omnih2o}. Recent work further improves data scalability through portable robot-in-the-loop or robot-free collection interfaces~\cite{ze2025twist2, nai2026humanoidmanipulationinterfacehumanoid}, models multiple action modes or improves generalization across objects, viewpoints, and scenes~\cite{qi2025coordinated, ze2025generalizable}, and leverages human demonstrations through single-video imitation via retargeting, joint human--humanoid training, or human-data pretraining followed by robot-data fine-tuning~\cite{li2024okami, qiu2025humanoid, heng2026humdex}. Together, these advances broaden the data sources and learning paradigms available for humanoid manipulation, spanning upper-body tasks and whole-body loco-manipulation.

Table~\ref{tab:comparisons} highlights a remaining gap. Existing humanoid manipulation systems support whole-body control with varying levels of end-effector dexterity and data-collection paradigms, but most do not incorporate tactile sensing or explicitly model tactile signals in the learned policy~\cite{he2024omnih2o, fu2024humanplus, lu2025mobile, li2025amo, ze2025twist2, luo2025sonic, nai2026humanoidmanipulationinterfacehumanoid, heng2026humdex}. Conversely, tactile-centric methods and datasets demonstrate the value of tactile information~\cite{heng2025vitacformer, kwon2025humanoid}, but do not provide a learned humanoid manipulation system that combines whole-body control, full end-effector dexterity, tactile sensing, and touch modeling. Our method targets this missing intersection through a single-stage touch-aware humanoid manipulation policy with implicit touch modeling via future-touch prediction.

\begin{figure*}
    \centering\includegraphics[width=1.0\linewidth]{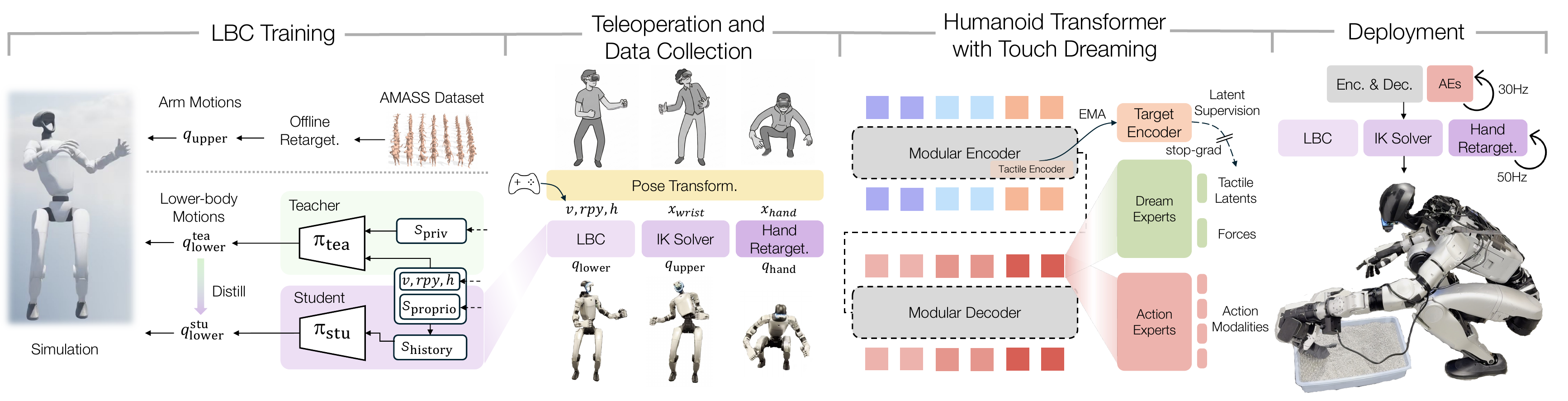}
    \caption{\textbf{System Overview.} \textbf{Left (LBC Training):} A teacher-student framework trains the lower-body controller (LBC) to track base velocity, torso orientation, and height, while robustly handling retargeted arm motions from the AMASS dataset. \textbf{Middle-Left (Teleoperation):} Human VR motions are mapped into unified torso commands (for LBC), end-effector poses (for IK), and hand targets (for retargeting), with a joystick dictating base velocity. \textbf{Middle-Right (Touch Dreaming):} A multimodal transformer policy processes vision, touch, and proprioception to predict action chunks alongside future hand joint forces and tactile latents. Future tactile latents are supervised by an EMA target encoder (teacher encoder in Sec.~\ref{sec:training}) with stop-gradient, providing stable latent targets. \textbf{Right (Deployment):} The policy streams action chunks at 30 Hz to the LBC, IK solver, and hand retargeter, all of which operate at 50 Hz.
}
    \label{fig:framework}
\end{figure*}

\subsection{Representation Learning for Contact-Rich Manipulation with Tactile Sensing}

Tactile sensing has increasingly been studied as a representation-learning problem rather than only a task-specific perception module. Early visuo-tactile manipulation works showed that touch complements vision for resolving contact state under partial observability~\cite{calandra2018more, lee2020making}. More recent work learns transferable tactile representations across sensors, tasks, and embodiments, improving data efficiency and reuse in downstream manipulation~\cite{higuera2024sparsh, zhao2024transferable}. In parallel, a growing line of visuo-tactile action models incorporates touch or force directly into policies for contact-rich manipulation, including diffusion-based, transformer-based, and VLA-style approaches~\cite{helmut2025tactile, xue2025reactive, heng2025vitacformer, huang20243d, zhu2025touch, chen2025multi, lin2025learning, yuan2026vtam, zheng2026omnivta, zhang2026dextac, chen2025implicitrdp, huang2025tactile, bi2025vla, zhang2025vtla}. These works consistently suggest that touch provides critical information about force, slip, compliance, and contact transitions that is difficult to infer from vision alone.

A closely related direction uses \emph{predictive} tactile learning to improve contact-aware representations. Prior work has explored self-supervised multimodal prediction for contact-rich tasks~\cite{lee2020making}, while more recent methods explicitly predict future tactile observations, tactile latents, or related contact quantities~\cite{heng2025vitacformer, ye2025learning, higuera2026visuo, zheng2026omnivta, yuan2026vtam, zhang2026dextac, chen2025implicitrdp, yoo2026aslipacousticsensingcontinuous}. These methods show that anticipating future touch can improve representation quality, planning, or reactive control. Some also rely on manually designed virtual targets tied to specific tactile sensor layouts~\cite{yuan2026vtam, chen2025implicitrdp}. Our method instead learns directly from future hand forces and EMA-supervised tactile latents, avoiding such sensor-specific target engineering. At the same time, much of this literature focuses on arm-hand manipulation and often relies on separate tactile pretraining, explicit world-model modules, or multi-stage inference in which predicted tactile signals are fed into a downstream policy or planner~\cite{zhao2024transferable, ye2025learning, zheng2026omnivta}.

In contrast, we use future-touch prediction not as a separate world model or inference-time module, but as an auxiliary objective inside a single-stage whole-body humanoid imitation policy. Our framework augments behavioral cloning with \emph{touch dreaming}: prediction of future hand forces together with future tactile latents supervised by an EMA teacher. This regularizes the shared Transformer trunk to learn contact-aware latent dynamics while keeping both training and deployment simple. Unlike prior work centered on arm-hand systems or multi-stage visuo-tactile pipelines, our method integrates future-touch prediction directly into a single-stage policy for dexterous, contact-rich \emph{whole-body humanoid} manipulation.

%% file: arxiv_contents/methodology.tex
\subsection{A System for Versatile Humanoid Dexterous Manipulation}

Fig.~\ref{fig:framework} presents our system for learning real-world, dexterous, contact-rich humanoid manipulation. The system consists of four stages: lower-body controller (LBC) training, VR-based teleoperation and data collection, policy learning with Humanoid Transformer with Touch Dreaming (HTD), and deployment. At its foundation is an RL-based LBC that provides stable lower-body and torso execution during manipulation. We train this controller in simulation with a teacher--student framework: a teacher policy learns robust lower-body behaviors under retargeted arm motions, and a deployable student policy imitates it using only proprioception and a short history. The resulting LBC tracks base velocity, torso orientation, and height commands, and serves as the execution backbone during both teleoperation and HTD policy deployment.

Building on this controller, we collect whole-body humanoid demonstrations through VR teleoperation. Human head, wrist, and hand motions are transformed into a unified robot reference frame and decomposed into torso commands for the LBC, end-effector pose targets for an IK solver, and hand targets for dexterous retargeting; the operator additionally provides base velocity commands through a joystick. The resulting dataset contains synchronized camera views, proprioception, hand-force signals, and tactile observations paired with whole-body action targets.

Using these demonstrations, we train HTD, a multimodal touch-aware loco-manipulation policy. HTD uses a modular encoder--decoder Transformer to tokenize multi-view images, robot and hand proprioception, hand-force signals, and tactile inputs into a shared latent representation, and to decode structured action outputs for the body and hands. In addition to action chunk prediction, HTD introduces touch-dreaming heads that predict future hand joint forces and future tactile latents. HTD is trained in a single stage with behavioral cloning augmented by these auxiliary touch-dreaming objectives. Future tactile latents are supervised by an EMA target encoder, which provides stable latent targets without requiring a separate tactile pretraining stage; gradients are stopped through the EMA encoder so it serves only as a slowly evolving target network. These auxiliary objectives regularize the shared Transformer trunk to learn contact-aware latent dynamics. During deployment, the policy streams action chunks to the LBC, IK solver, and hand retargeter, while the dream heads are used only during training and are not executed at inference time.

\subsection{Lower-body Controller}
We train the humanoid lower-body policy in massively parallel simulation with Isaac Lab~\cite{mittal2025isaac}.
The lower-body policy is command-conditioned and aims to track base motion and torso pose targets.
At each control step $t$, the deployable proprioceptive observation $\bm{s}_{\mathrm{proprio}}^{t}$ is defined as
\begin{equation}
\bm{s}_{\mathrm{proprio}}^{t}
=
\big[
\bm{\omega}^{t},\;
\bm{g}^{t},\;
\bm{q}_{\mathrm{lower}}^{t},\;
\dot{\bm{q}}_{\mathrm{lower}}^{t},\;
\bm{a}_{\mathrm{lower}}^{t-1}
\big].
\end{equation}



Here, $\bm{\omega}^{t}$ is the base angular velocity and $\bm{g}^{t}$ denotes the projected gravity vector, both expressed in the body frame; $\bm{q}_{\mathrm{lower}}^{t}$ and $\dot{\bm{q}}_{\mathrm{lower}}^{t}$ are the lower-body joint positions and velocities, and $\bm{a}_{\mathrm{lower}}^{t-1}$ is the previous lower-body action. The action output $\bm{q}_{\mathrm{lower}} \in \mathbb{R}^{15}$ is a 15-dimensional vector of target joint positions: $2 \times 6$ for the two legs and $3$ for the waist motors.

We adopt a teacher--student framework to train the lower-body policy. The teacher policy is first trained in simulation using PPO~\cite{schulman2017proximal} with access to privileged information, and is subsequently distilled into a student policy via DAgger~\cite{ross2011reduction}. The student policy observes only information available in the real world and can therefore be deployed for both teleoperation and autonomous execution. During training, the upper-body joints are not controlled by this policy; instead, we replay retargeted arm joint references sampled from AMASS~\cite{mahmood2019amass} to simulate the torques and disturbances induced by upper-body manipulation.


\parab{Teacher policy.} The teacher policy is formulated as
\begin{equation}
\pi^{T}\!\left(
\bm{s}_{\mathrm{proprio}},\;
\bm{s}_{\mathrm{priv}},\;
\bm{v},\;
\bm{rpy},\;
h
\right)
=
\bm{q}_{\mathrm{lower}}^{T},
\end{equation}
where $\bm{q}_{\mathrm{lower}}^{T} \in \mathbb{R}^{15}$ represents the 15-DoF lower-body target joint positions. The privileged observation $\bm{s}_{\mathrm{priv}} = \bm{c}_{\mathrm{feet}}$ is a binary foot-contact indicator $\bm{c}_{\mathrm{feet}} \in \{0,1\}^{2}$ available in simulation. The command inputs include the base velocity $\bm{v}$, the torso orientation $\bm{rpy}$, and the torso height $h$. The teacher policy maximizes a weighted sum of tracking rewards for commanded base motion and torso pose, together with regularization, contact, gait, stability, and termination terms.

\begin{figure}[tp]
    \centering
    \includegraphics[width=1.0\linewidth]{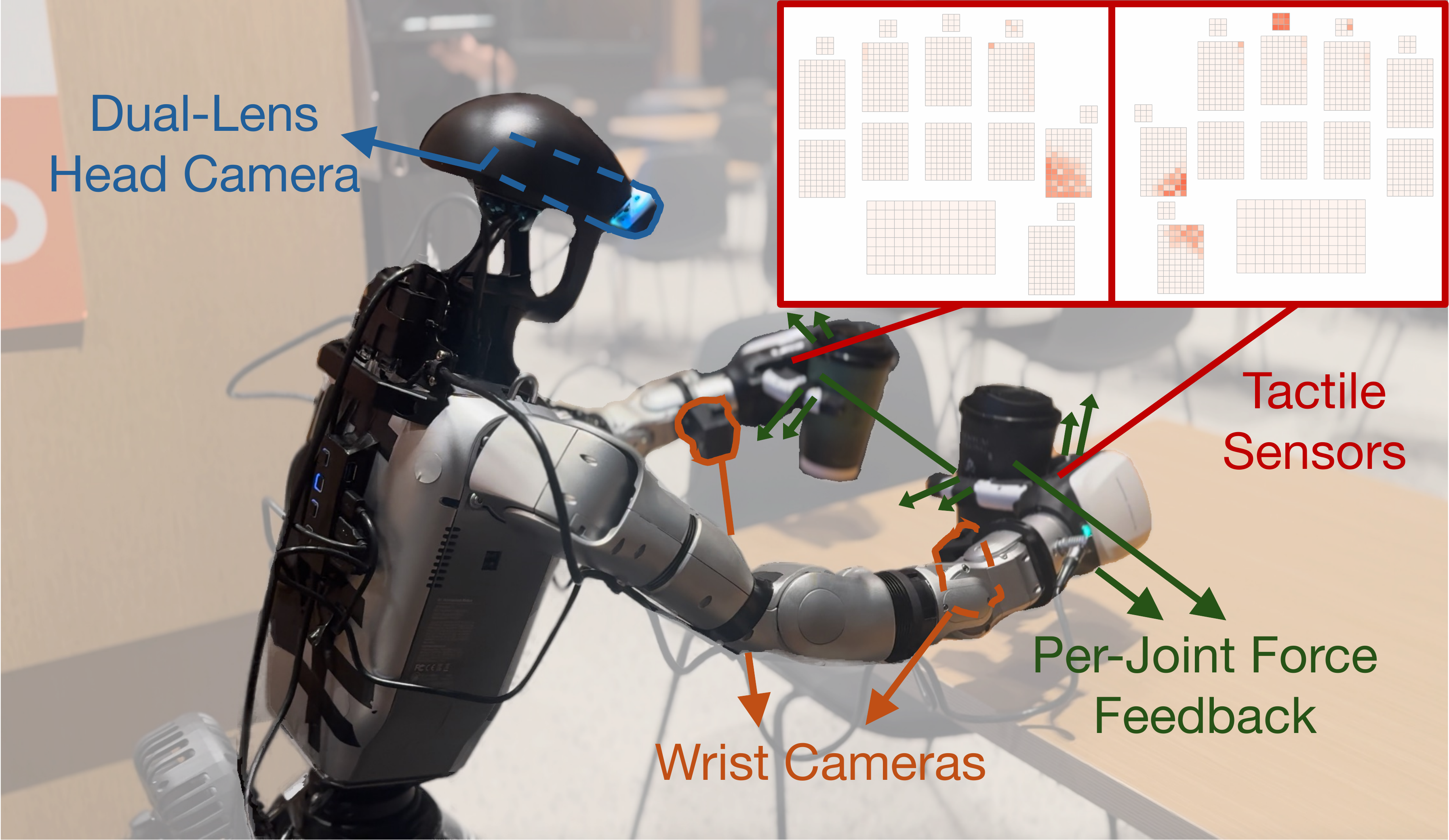}
    \caption{\textbf{System setup.} Hardware used for whole-body humanoid data collection and policy learning, including a dual-lens head camera, wrist cameras, dexterous hands equipped with distributed tactile sensors, and per-joint force feedback from the hand joints. The tactile layout covers the fingers and palms of both hands, and the inset visualizes the corresponding sensor maps together with representative contact activations.}
    \label{fig:system_setup}
\end{figure}

\begin{figure*}[htp]
    \centering
    \includegraphics[width=1.0\linewidth]{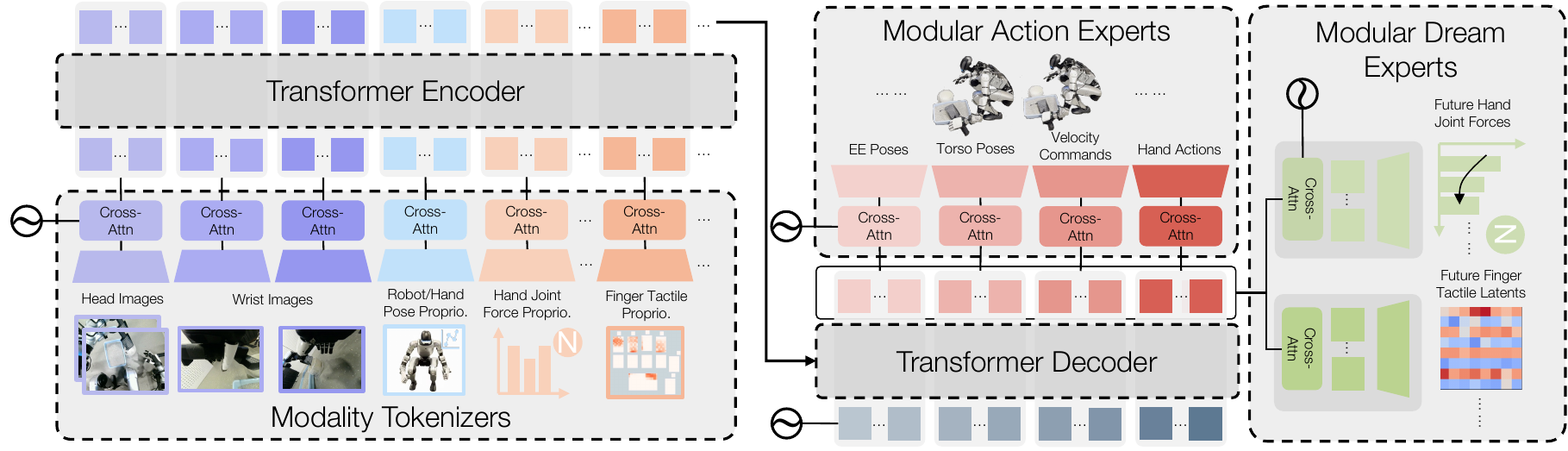}
    \caption{\textbf{HTD model architecture.} HTD is a modular encoder--decoder Transformer. \textbf{Left:} modality tokenizers encode multi-view images, proprioception, hand joint forces, and tactile signals into a fixed number of tokens via cross-attention aggregation. \textbf{Middle:} a Transformer encoder fuses multimodal observation tokens, and a Transformer decoder produces a fixed set of output tokens. \textbf{Right:} modular \emph{action experts} decode pose/velocity/hand-action targets, while modular \emph{dream experts} predict future forces and tactile latents for touch dreaming.  We use learnable query embeddings to flexibly determine how many tokens to use for each input/output modality.}
    \label{fig:arch}
\end{figure*}


\parab{Student policy.} We distill the teacher into a deployable student policy via DAgger. The student policy consumes only real-world-available observations:
\begin{equation}
\pi^{S}\!\left(
\bm{s}_{\mathrm{proprio}},\;
\bm{s}_{\mathrm{history}},\;
\bm{v},\;
\bm{rpy},\;
h
\right)
=
\bm{q}_{\mathrm{lower}}^{S}.
\end{equation}
To compensate for partial observability, the student concatenates a 2-timestep history of proprioceptive observations $\bm{s}_{\mathrm{history}}$. During training, the student rolls out its own actions in simulation while being supervised by the teacher's reference actions at each timestep, minimizing the $L_2$ loss $\mathcal{L} = \|\bm{q}_{\mathrm{lower}}^{S} - \bm{q}_{\mathrm{lower}}^{T}\|_2^2$ between student and teacher outputs.

\parab{Training details.}
During training, command signals are uniformly sampled from predefined ranges to cover diverse loco-manipulation behaviors. We also apply domain randomization to improve sim-to-real transferability. Additional details of LBC training are provided in Appendix~\ref{sec:appendix_lbc}.


\subsection{Teleoperation and Data Collection}

As summarized in Fig.~\ref{fig:framework}, our demonstration pipeline couples VR-based motion mapping with whole-body command execution to collect synchronized humanoid trajectories in real-world settings. At runtime, the operator's head, wrist, and hand motions are transformed from the VR frame into a unified robot reference frame. From these signals, we derive torso pose commands $(\bm{rpy}, h)$, 6D wrist pose targets $\bm{x}_{\mathrm{wrist}}$ for upper-body execution, and hand targets $\bm{x}_{\mathrm{hand}}$ for dexterous retargeting. The base velocity command $\bm{v}$ is provided separately through a joystick. This design lets the operator focus on task intent and dexterous interaction, while the robot-side control stack handles stabilization and low-level execution.

These targets are executed through a three-stage stack. First, the LBC takes $(\bm{v}, \bm{rpy}, h)$ and produces lower-body joint targets $\bm{q}_{\mathrm{lower}}$ to maintain stable locomotion, posture, and torso tracking. Second, an IK solver maps the desired wrist/end-effector poses $\bm{x}_{\mathrm{wrist}}$ to upper-body joint targets $\bm{q}_{\mathrm{upper}}$. Third, a hand retargeting module based on DexPilot~\cite{handa2020dexpilot} converts the human hand targets $\bm{x}_{\mathrm{hand}}$ into dexterous hand joint targets $\bm{q}_{\mathrm{hand}}$ by optimizing fingertip-distance consistency for reliable grasping and in-hand interaction. Together, this stack enables coordinated whole-body teleoperation while preserving full end-effector dexterity.


During teleoperation, we record synchronized multimodal observations from the humanoid, including RGB images from the dual-lens head camera and wrist cameras, robot and hand proprioception, per-joint force feedback from the dexterous hands, and tactile readings from both hands. Each hand provides a 1062-dimensional tactile observation distributed across 17 spatial sensing regions spanning the finger segments and palm surfaces. This distributed tactile layout captures localized contact patterns over the hand surface, as visualized in Fig.~\ref{fig:system_setup}. The resulting dataset pairs whole-body action commands ($\bm{v}, \bm{rpy}, h$, $\bm{x}_{\mathrm{wrist}}$, and $\bm{x}_{\mathrm{hand}}$) with multi-view vision, robot and hand proprioception, per-joint hand-force feedback, and distributed tactile observations for downstream policy learning.

\subsection{Learning Dexterous Manipulation with Touch Dreaming}
\label{sec:htd}

We aim to learn a versatile humanoid manipulation policy that robustly handles contact-rich interactions by modeling \emph{touch} as a core modality.
We introduce \textbf{Humanoid Transformer with Touch Dreaming (HTD)}, shown in Fig.~\ref{fig:arch}.
HTD follows a modular design with three groups of components: (i) modality tokenizers that encode each observation stream into tokens, (ii) an encoder--decoder transformer trunk that fuses multimodal information and models complex dynamics, and (iii) modular experts that decode the trunk outputs into both control actions and auxiliary touch-dreaming predictions.
Concretely, given observations including multi-view vision, robot and hand proprioception, hand joint force signals, and tactile readings, the tokenizers jointly produce a sequence of tokens that is fused by the transformer encoder.
The transformer decoder then emits a fixed set of output tokens, with each action modality assigned a fixed number of tokens. These tokens are consumed by two families of heads: \emph{action experts} that predict structured action targets for whole-body control, and \emph{dream experts} that predict future touch signals (forces and tactile latents) for touch dreaming. The \emph{dream experts} attend to the full set of output tokens across all action modalities. This design incentivizes the latent dynamics of the shared transformer trunk to be contact-aware. 

\parab{Modality Tokenizers.}
Each tokenizer $T_m$ maps a raw modality into a fixed number of tokens, which are concatenated in a fixed order to form the transformer encoder input, similar to \cite{niu2025human2locoman, wang2024scaling}.
As illustrated in Fig.~\ref{fig:arch} (left), we first extract modality-specific features, then compress them into tokens using a cross-attention aggregation layer, where a small set of learnable query (``slot'') tokens attends to the feature sequence.
For image modalities, we extract features with a pretrained ResNet~\cite{he2016deep} backbone (finetuned during training) and use separate tokenizers for the head camera and each wrist camera.
For state-like modalities (e.g., robot/hand pose and proprioception, as well as force-related proprioceptive signals), we use lightweight MLP feature extractors.
For tactile inputs, we use a dedicated tactile encoder to embed the raw tactile readings into a compact feature sequence, which is then tokenized via the same cross-attention aggregation.

\parab{Per-Finger/Region Tactile Encoder.}
For tactile inputs, we encode each finger or hand region independently rather than forming a single full-hand tactile embedding upfront. Concretely, the tactile observations are decomposed into anatomically defined inputs corresponding to the thumb, index finger, middle finger, ring finger, pinky, and palm. For a non-thumb finger, the 185-dimensional tactile input is further segmented into three local patches (tip, top, and palm-facing region); for the thumb, the 210-dimensional input is segmented into four patches (tip, top, mid, and palm-facing region); and for the palm, the 112-dimensional input is treated as a single large patch. Each local patch is reshaped into a 2D map and processed by a dedicated CNN branch selected according to patch size, with lightweight single-layer convolutions for small patches and deeper two-layer CNN blocks for larger patches. The resulting patch features are adaptively pooled to a fixed spatial resolution, flattened, concatenated, and fused by an MLP into a compact embedding for that finger or region. These per-region embeddings are then projected to the Transformer hidden dimension and converted into tactile tokens through the same cross-attention aggregation used by the other modality tokenizers. The same per-region tactile encoder architecture is also used to instantiate the EMA target encoder for stable latent supervision during touch dreaming.

\parab{Transformer Trunk.}
HTD uses an encoder--decoder transformer trunk with fixed input and output sequence lengths determined by the number of tokens allocated to each modality and each output group.
The encoder contextualizes the concatenated observation tokens into a unified representation.
The decoder produces a fixed set of output tokens at pre-specified positions. The learnable query embeddings serve as a structured interface that supports multiple downstream experts.
This separation enables the encoder to focus on multimodal state understanding while the decoder provides disentangled readouts for control and prediction.

\parab{Modular Action Experts.}
We decode control outputs with a set of \emph{modular action experts} (Fig.~\ref{fig:arch}, top-right). Each expert uses a cross-attention layer to read from the decoder output tokens and predicts a particular action modality, including end-effector pose targets, torso pose targets, velocity commands (when applicable), and hand actions. This modular design allows action modalities with different dimensionalities and control roles to be read out independently and adaptively. In particular, each action modality is assigned its own fixed number of decoder output tokens, so low-dimensional but behaviorally important outputs such as velocity commands can still receive sufficient representational capacity, while higher-dimensional outputs such as pose or hand-action targets can be decoded by separate experts matched to their complexity. We adopt action chunking~\cite{zhao2023learningfinegrainedbimanualmanipulation}, where each expert predicts a short horizon of targets at each inference step.

\parab{Modular Dream Experts and Touch Dreaming.}
In addition to action experts, HTD includes \emph{modular dream experts} that provide auxiliary prediction objectives during training (Fig.~\ref{fig:arch}, far-right).
These experts predict future touch outcomes, including (i) future hand joint force vectors and (ii) future finger/region tactile \emph{latents}.
We refer to these auxiliary predictions as \textit{touch dreaming}: conditioned on the current multimodal observations, the model ``imagines'' near-future touch feedback, which regularizes the shared transformer trunk to learn contact-aware representations.
Crucially, for tactile we perform prediction in a learned latent space rather than raw sensor space.
Direct regression in raw tactile space is often dominated by sparsity and noise, whereas latent supervision provides a compact target that captures contact structure.
We obtain stable latent labels using an EMA tactile tokenizer as a teacher (Sec.~\ref{sec:training}), and supervise the student to match these teacher latents.
During deployment, only the action experts are used for control; dream experts' outputs are not used.

\parab{Modality Decomposition.}
We preserve semantically distinct inputs as separate modalities and tokenize them independently.
On the output side, we similarly decode different action modalities with separate action experts, and decode touch-dreaming targets with dedicated dream experts. Given the distinct statistics of different input/output modalities, this network design allows for modality-based specialization, while the shared transformer trunk learns a unified representation and models complex dynamics. 

\begin{algorithm}[t]
\caption{Imitation learning with touch dreaming}\label{algo:train_touch_dream_ema}
\begin{algorithmic}[1]
\REQUIRE Dataset $\mathcal{D}$ with tuples $(\bm{o}_t, A_t, F_t, S_t)$
\REQUIRE Shared action and touch prediction horizon $H$
\REQUIRE EMA decay $\alpha$, loss weights $\lambda_F,\lambda_Z$, magnitude weight $\beta$, learning rate $\eta$
\ENSURE HTD Policy $\pi_{\Theta}$ 
\STATE Initialize policy $\pi_{\Theta}$ ($\Theta$ has parameters from tokenizers, encoder--decoder trunk, and detokenizers; $\theta\in \Theta$)
\STATE Initialize teacher tactile tokenizer, $\theta^T=\theta$
\FOR{step $=1,2,\ldots$}
    \STATE Sample a batch $B=\{(\bm{o}_j, A_j, F_j, S_j)\}_{j=1}^{n}$ from $\mathcal{D}$
    \STATE \textbf{Teacher latents:} compute $\bm{z}^{\star}_{j,k}$ with Eq.~\eqref{eq:teacher_latents}
    \STATE \textbf{Policy rollout:} $ \hat{A}_j, \{\hat{
    \bm{f}}_{j,k}\}_{k=1}^{H}, \{\hat{\bm{z}}_{j,k}\}_{k=1}^{H} \sim  \pi_{\Theta}(\bm{o}_j)$
    \STATE Compute total loss $\mathcal{L}(B;\Theta)$ with Eq.~\eqref{eq:total_loss}
    \STATE Update student: $\Theta^S \leftarrow \Theta^S - \eta \nabla_{\Theta^S} \mathcal{L}(B; \Theta)$
    \STATE Update EMA teacher: $\theta^T \leftarrow \alpha \theta^T + (1-\alpha)\theta^S$ 
\ENDFOR
\RETURN $\pi_{\Theta}$
\end{algorithmic}
\end{algorithm}

\subsection{Training Paradigm}
\label{sec:training}
We train the policy with a single-stage behavioral cloning (BC) paradigm on humanoid demonstrations.
The key component of our architecture is \textit{touch dreaming}: in addition to predicting action chunks, the model is trained to predict future touch signals.
Specifically, we (i)~predict \emph{future hand joint force vectors} using a smooth L1 loss, and (ii)~predict \emph{future tactile latents} in a stable latent space generated by an EMA teacher encoder.
We find that supervising tactile predictions in latent space instead of regressing raw tactile arrays yields substantially better manipulation performance on real robots, because it provides a compact and semantically rich learning signal while avoiding the difficulty of reconstructing sparse, high-dimensional sensor readings.
This auxiliary predictive objective encourages the Transformer trunk to learn contact-aware world representations that improve downstream contact-rich manipulation. The overall training procedure is summarized in Algorithm~\ref{algo:train_touch_dream_ema}.

\parab{EMA Teacher for Tactile Latents.}
Let $T_{\rm tact}(\theta)$ denote the student tactile tokenizer parameterized by $\theta$ and $T^{T}_{\rm tact}(\theta^T)$ its EMA counterpart.
After each optimization step, teacher parameters are updated as an EMA of the student parameters:
\begin{align}
\theta^{T} \leftarrow \alpha \theta^{T} + (1-\alpha)\theta, \qquad \alpha \in (0,1),
\label{eq:ema}
\end{align}
and no gradient is backpropagated through the teacher.
The teacher network provides slowly evolving, temporally consistent latent targets.
Without the stop-gradient EMA target, the tactile tokenizer and touch detokenizer can co-adapt toward near-constant latent representations, making the tactile-prediction objective uninformative.

\parab{Objective.}
Let the dataset
$\mathcal{D}=\{(\bm{o}_t, A_t, F_t, S_t)\}$,
where $\bm{o}_t$ is the multimodal observation at time $t$,
$A_t=\{\bm{a}_{t+\ell}\}_{\ell=1}^{H}$ is the action chunk of horizon $H$,
$F_{t}=\{\bm{f}_{t+\ell}\}_{\ell=1}^{H}$ is the future hand joint force sequence,
and $S_{t}=\{\bm{s}_{t+\ell}\}_{\ell=1}^{H}$ is the future tactile signal sequence; all three target sequences share the horizon $H$.
Given action modalities $m_1,\ldots,m_K$ and touch signals (force and tactile), the overall loss is: %
\begin{align}
\mathcal{L}(\Theta)
= \underbrace{\sum_{i=1}^{K} \mathcal{L}_{\rm act,\, m_i}(\Theta)}_{\text{behavior cloning}}
+ \lambda_F \underbrace{\mathcal{L}_{\rm force}(\Theta)}_{\text{force pred.}}
+ \lambda_Z \underbrace{\mathcal{L}_{\rm tact}(\Theta)}_{\text{tactile latent pred.}},
\label{eq:total_loss}
\end{align} %
where $\lambda_F$ and $\lambda_Z$ weight the touch-related objectives. We denote the smooth L1 losses for action, force, and latent-magnitude regression by $\ell_{\rm act}$, $\ell_{\rm force}$, and $\ell_{\rm mag}$, respectively. For a batch $B=\{(\bm{o}_j, A_j, F_j, S_j)\}_{j=1}^{n}$, the BC loss with action chunking is:
\begin{align}
\mathcal{L}_{\rm act,\, m_i}(B;\Theta)
= \frac{1}{n}\sum_{j=1}^{n}\Bigg[\frac{1}{H}\sum_{\ell=1}^{H}
\ell_{\rm act}\!\left(\bm{a}_{j,\ell}[m_i], \hat{\bm{a}}_{j,\ell}[m_i]\right)\Bigg],
\label{eq:bc_loss}
\end{align}%
where $\hat{\bm{a}}_{j,\ell} = [\pi_\Theta(\bm{o}_j)]_{\ell}$ denotes the $\ell$-th action in the chunk.

\parab{Future Hand Joint Force Prediction Loss.}
For force dreaming, the model predicts future force vectors $\hat{\bm{f}}_{j,k}$ for $k \in \{1,\ldots,H\}$, supervised with the same smooth L1 loss used for action prediction:
\begin{align}
\mathcal{L}_{\rm force}(B;\Theta)
= \frac{1}{n}\sum_{j=1}^{n}\Bigg[\frac{1}{H}\sum_{k=1}^{H}
\ell_{\rm force}\!\left(\hat{\bm{f}}_{j,k},\, \bm{f}_{j,k}\right)\Bigg].
\label{eq:force_loss}
\end{align}

\parab{Tactile Dreaming Loss (Latent Supervision).}
For tactile dreaming, we supervise the model to predict \emph{future tactile latents} rather than raw tactile heatmaps.
For each future step $k \in \{1,\ldots,H\}$, we compute target latent labels by encoding future tactile measurements with the EMA teacher encoder:
\begin{align}
\bm{z}^{\star}_{j,k} &= \mathrm{stopgrad}\!\left(T^{T}_{\rm tact}(\bm{s}_{j,k})\right),
\label{eq:teacher_latents}
\end{align}
and the touch detokenizer predicts $\hat{\bm{z}}_{j,k}$.
We combine a cosine direction loss with a magnitude alignment loss:
\begin{align}
\mathcal{L}_{\rm tact}(B;\Theta)
&= \frac{1}{n}\sum_{j=1}^{n}\Bigg[\frac{1}{H}\sum_{k=1}^{H}
\Big(\underbrace{1 - \cos(\hat{\bm{z}}_{j,k},\, \bm{z}^{\star}_{j,k})}_{\text{direction}} \notag\\
&\quad + \beta\,\underbrace{\ell_{\rm mag}\!\left(\|\hat{\bm{z}}_{j,k}\| - \|\bm{z}^{\star}_{j,k}\|\right)}_{\text{magnitude}}\Big)\Bigg],
\label{eq:tact_latent_loss}
\end{align}
where $\cos(\cdot,\cdot)$ denotes cosine similarity and $\beta$ weights magnitude alignment.
The direction term aligns the predicted latent with the teacher target in orientation, while the magnitude term matches their norms, preventing collapse to unit-norm predictions that satisfy cosine similarity alone.

%% file: arxiv_contents/experiments.tex
We conduct experiments to answer the following research questions: 
\begin{enumerate*}
    \item How does our LBC strategy compare to other methods with regard to tracking accuracy and robustness?
    \item How do our learning system and HTD perform on real-world versatile humanoid manipulation?
    \item How do predictive touch dreaming and latent tactile supervision impact the effectiveness of contact-rich humanoid manipulation?
    \item Does touch dreaming capture meaningful and robust contact-aware representations?
\end{enumerate*}

\subsection{How does our LBC strategy compare to other methods with regard to tracking accuracy and robustness?}
To evaluate tracking performance, we benchmark our LBC against two representative decoupled whole-body humanoid controllers: FALCON~\cite{zhang2025falcon}, which uses a dual-policy architecture with separate upper- and lower-body control and adaptive force curriculum learning; and AMO~\cite{li2025amo}, which uses a hierarchical whole-body control framework where upper-body goals and torso commands are converted into lower-body references through an adaptive motion optimization module and tracked by a learned lower policy.
We quantify performance with tracking error metrics:
\begin{itemize}
    \item Linear Velocity Tracking Error $E_v$: measures the L2 error between commanded and actual forward/lateral velocities in the robot's yaw-aligned horizontal frame.
    \item Angular Velocity Tracking Error $E_\omega$: captures the absolute difference between commanded and actual yaw angular velocity in the world frame.
    \item Height Tracking Error $E_h$: quantifies the deviation of torso height from the commanded value.
    \item Yaw Orientation Tracking Error $E_y$: evaluates the error in relative yaw angle between the torso and the pelvis.
    \item Pitch Orientation Tracking Error $E_p$: measures the absolute pitch angle deviation of the torso from the commanded upright orientation.
    \item Roll Orientation Tracking Error $E_r$: assesses the error in relative roll angle between torso and pelvis.
\end{itemize}
All orientations are computed using intrinsic XYZ Euler decomposition. For evaluation, we run each baseline using its publicly available implementations across 4096 parallel simulation environments for 500 timesteps. We calculate metrics by averaging tracking errors over all timesteps and environments, providing a statistically robust assessment of sustained tracking performance across varied conditions.

\begin{figure}[t]
    \centering
    \includegraphics[width=1.0\linewidth]{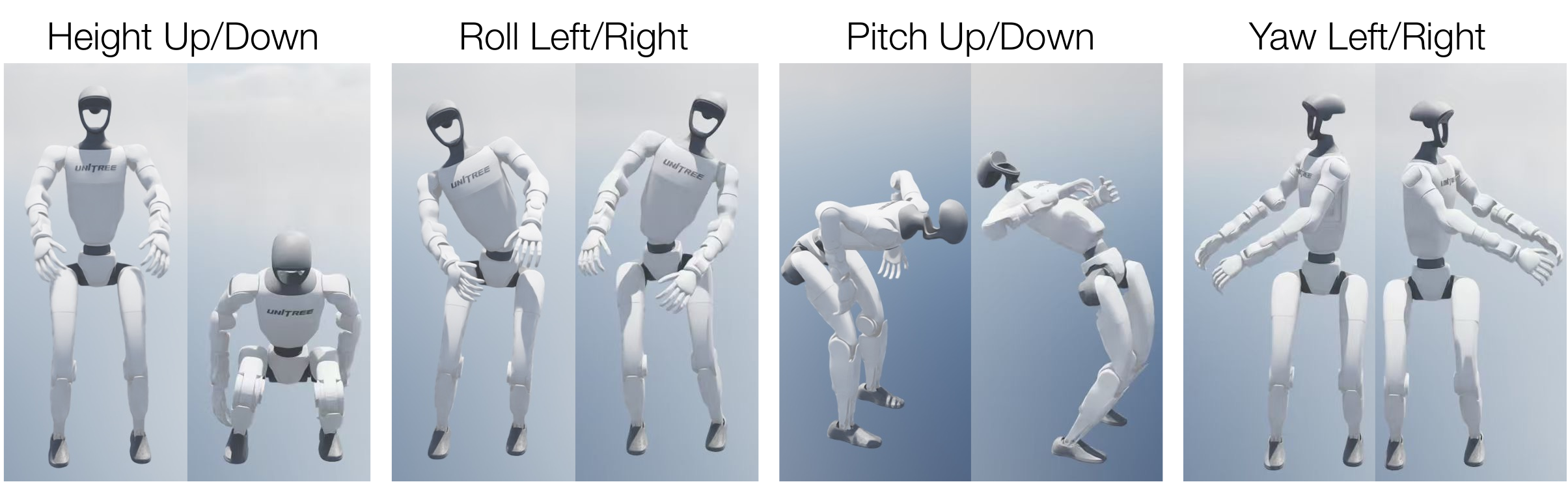}
    \caption{Visualization of postures near the boundary of the stable controllable workspace of our LBC policy in simulation.}
    \label{fig:wbc_range}
\end{figure}

Our LBC achieves the best overall tracking on most metrics in Table~\ref{tab:tracking_error}, notably linear velocity, torso height, and torso orientation ($E_v$, $E_h$, $E_y$, $E_p$, and $E_r$), indicating tighter torso pose regulation that is crucial for contact-rich whole-body manipulation. While AMO attains a slightly lower yaw rate error $E_\omega$, our controller provides a better balance between locomotion tracking and whole-body configuration control. We note that the standard deviations of $E_h$ and especially $E_p$ exceed their means, largely due to a small number of difficult command combinations that create conflicting objectives, e.g., simultaneously requesting large forward torso pitch and a low torso height; in these cases, the policy may temporarily trade off tracking for stability, increasing variance despite low average errors across the command space.

\begin{table}[h]
\centering
\caption{Tracking Error Comparison}
\vspace{-0.2cm}
\label{tab:tracking_error}
\resizebox{\columnwidth}{!}{%
\begin{tabular}{l|ccc}
\toprule
\textbf{Metric} & \textbf{Ours} & \textbf{AMO~\cite{li2025amo}} & \textbf{FALCON~\cite{zhang2025falcon}} \\
\midrule
$E_v$ (m/s) & \textbf{0.1420 $\pm$ 0.0568} & 0.1779 $\pm$ 0.0642 & 0.1641 $\pm$ 0.0309 \\
$E_\omega$ (rad/s) & 0.1806 $\pm$ 0.0534 & \textbf{0.1540 $\pm$ 0.0316} & 0.1874 $\pm$ 0.0263 \\
$E_h$ (m) & \textbf{0.0280 $\pm$ 0.0438} & 0.0568 $\pm$ 0.0814 & 0.1299 $\pm$ 0.0082 \\
$E_y$ (rad) & \textbf{0.0126 $\pm$ 0.0051} & 0.1540 $\pm$ 0.0534 & 0.1215 $\pm$ 0.0111 \\
$E_p$ (rad) & \textbf{0.0487 $\pm$ 0.1796} & 0.1519 $\pm$ 0.1254 & \textcolor{gray}{\emph{(not tracked)}} \\
$E_r$ (rad) & \textbf{0.0157 $\pm$ 0.0065} & 0.0735 $\pm$ 0.0447 & \textcolor{gray}{\emph{(not tracked)}} \\
\bottomrule
\end{tabular}
}
\end{table}

Average tracking error alone does not fully characterize the usable operating envelope of a whole-body controller, since manipulation-oriented control requires stability under large torso reorientation and height variation that expand reachability. We therefore measure the per-dimension stable controllable range of our LBC policy in simulation by sweeping one command dimension at a time, torso height $h$, torso roll $\phi_{\text{torso}}$, torso pitch $\theta_{\text{torso}}$, and torso yaw $\psi_{\text{torso}}$, while keeping the others nominal, and running the policy until instability, a fall, or persistent tracking failure occurs; we report the largest interval that remains stable over the rollout. 
Our policy achieves stable command-tracking ranges of $h \in [0.33,\,0.80]$\,m, $\phi_{\text{torso}} \in [-0.38,\,0.35]$\,rad, $\theta_{\text{torso}} \in [-0.92,\,1.41]$\,rad, and $\psi_{\text{torso}} \in [-1.50,\,1.34]$\,rad. The stable workspace covers the trained height range and most of the yaw range, while roll remains the most restrictive.
Fig.~\ref{fig:wbc_range} visualizes boundary postures, showing that our controller supports a wide stable region for crouching, bending, and large torso reorientation beyond nominal upright motions.

\subsection{How do our learning system and HTD perform on real-world versatile humanoid manipulation?}
To answer this question, we evaluate our learning system on five humanoid manipulation tasks (as shown in Fig.~\ref{fig:head}):
\begin{itemize}
    \item \textbf{Insert-T:} The robot grasps a T-shaped block randomly initialized within a region on the table and inserts it into a fixed T-shaped base with a tight clearance of $3.5$\,mm. This task tests contact-sensitive correction and high spatial accuracy under small insertion tolerances.
    
    \item \textbf{Book Organization:} The robot gently pushes a hardcover book, chosen from two variants and randomly initialized within a region on the table, to create a graspable overhang, since the book is difficult to pick up directly from a flat surface. It then securely grasps the book and places it onto a bookshelf. This task tests hybrid pushing-and-grasping and controlled reorientation and placement for thin rigid objects with limited grasp affordance.
    
    \item \textbf{Towel Folding:} The robot folds a towel on the table. The towel is randomly initialized within a region on the table using one of three types of initial configurations. This task tests deformable object handling over multiple manipulation stages.
    
    \item \textbf{Cat Litter Scooping:} The robot squats to reach a litter scoop on the ground, uses it to scoop 3D-printed litter from a box, and dumps the litter into a trash bin. The litter is randomly distributed within the box, the trash bin is randomly placed on the right side of the box, and the scoop is randomly placed near the left edge of the box with two pose variations. This task tests tool-mediated interaction and whole-body reachability.
    
    \item \textbf{Tea Serving:} The robot walks to a bar, picks up two cups of tea randomly positioned within a region on the bar table, carries them to a nearby table, comes to a stop, and places both cups on the tabletop. This task tests dual-arm loco-manipulation while maintaining object stability throughout the motion.
\end{itemize}
\parab{Training demonstrations.}
The task-specific datasets used for policy training contain 193, 150, 141, 153, and 72 trajectories for \textit{Insert-T}, \textit{Book Organization}, \textit{Towel Folding}, \textit{Cat Litter Scooping}, and \textit{Tea Serving}, respectively. These trajectories correspond to approximately one hour of demonstrations per task and about 40 minutes for \textit{Tea Serving}.

\parab{Baselines.}
We compare our approach against two decoder-only variants of ACT~\cite{zhao2023learningfinegrainedbimanualmanipulation}, which we found empirically to perform better than the version augmented with a CVAE encoder. \textbf{ACT (Visual + Proprio)} uses only multi-view vision and proprioception. \textbf{ACT (Visual + Proprio + Touch)} additionally takes force and tactile observations as input. We report results for our method, \textbf{HTD}, which augments imitation learning with touch dreaming and employs a modular design for observations, action experts, and dream experts.

\parab{Metrics and protocol.}
For each task and method, we run 20 real-world trials and report the task score (mean $\pm$ SEM) and success rate. 
Task score reflects task completion quality under partial progress; success rate measures strict task completion.

\begin{figure}[!t]
    \centering
    \includegraphics[width=1.0\linewidth]{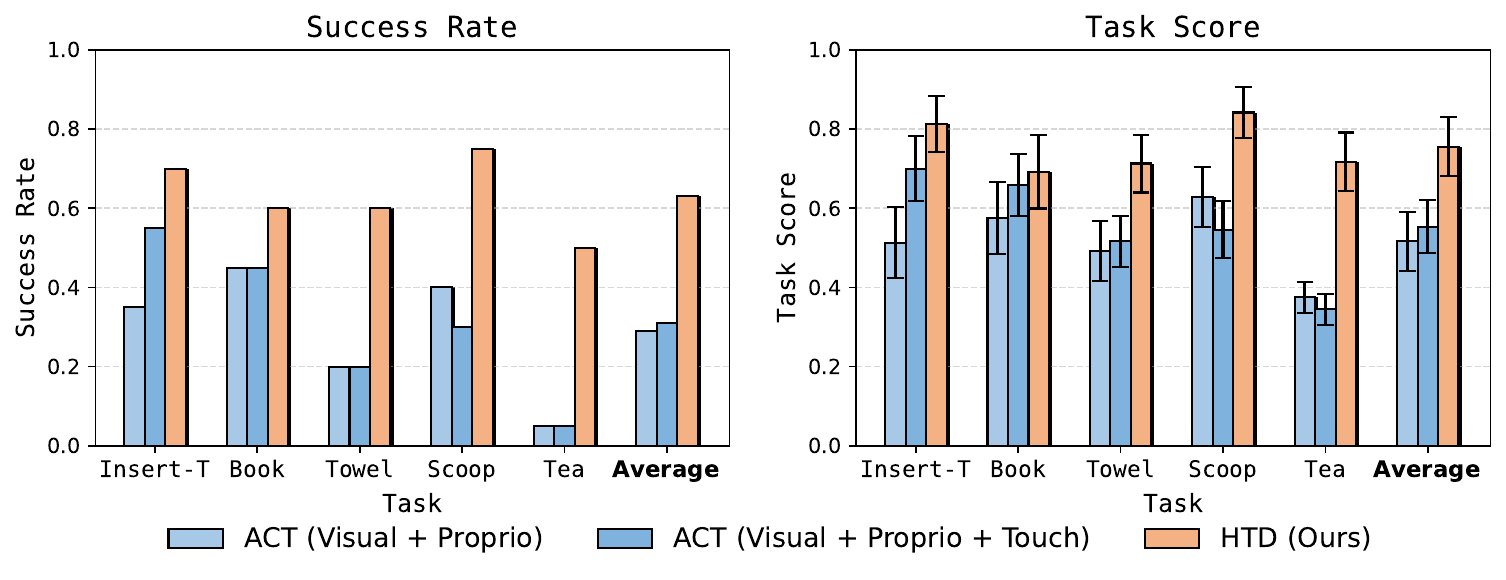}
    \caption{
Real-world results on five contact-rich tasks. We compare ACT (Visual + Proprio), ACT (Visual + Proprio + Touch), and HTD. Left: success rate. Right: task score (mean $\pm$ SEM, 20 trials).
    }
    \label{fig:main_results}
\end{figure}
\begin{figure}[!t]
    \centering
    \includegraphics[width=1.0\linewidth]{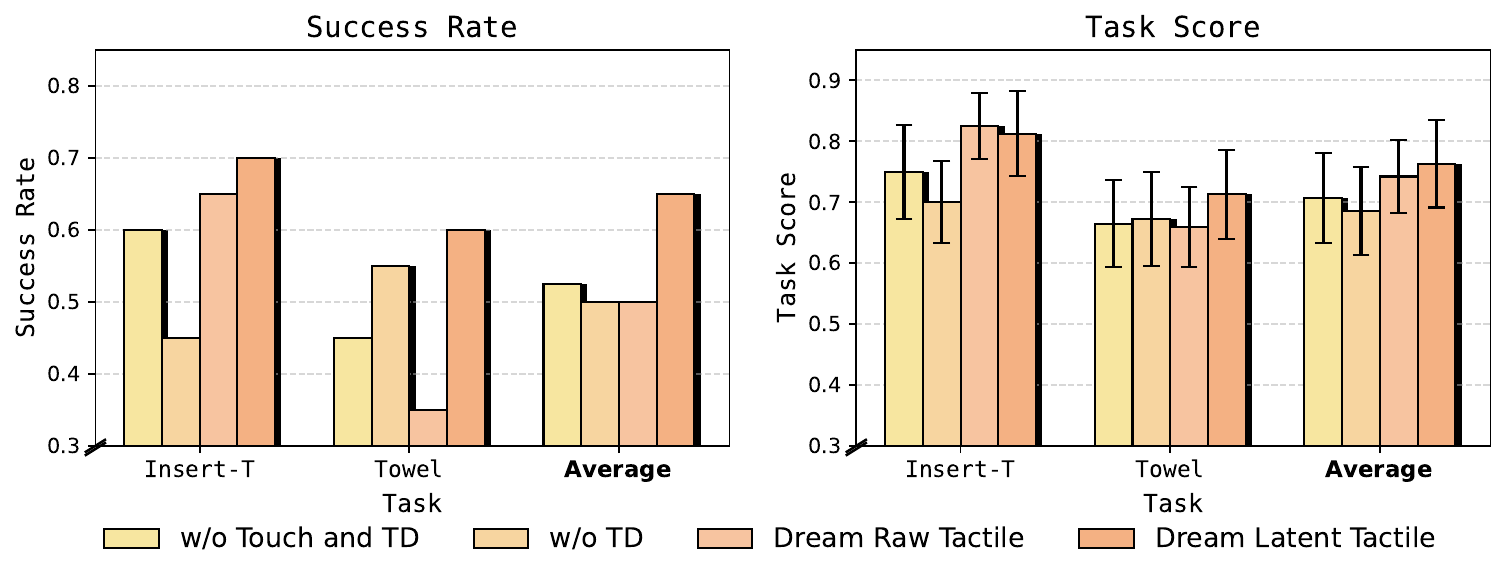}
    \caption{
Ablations of HTD. Variants: w/o Touch and TD, w/o TD, Dream Raw Tactile, and Dream Latent Tactile (full). Left: success rate. Right: task score (mean $\pm$ SEM, 20 trials).
    }
    \label{fig:ablation_results}
\end{figure}

\begin{figure*}[!t]
    \centering
    
    \begin{subfigure}{1.0\linewidth}
        \includegraphics[width=\linewidth]{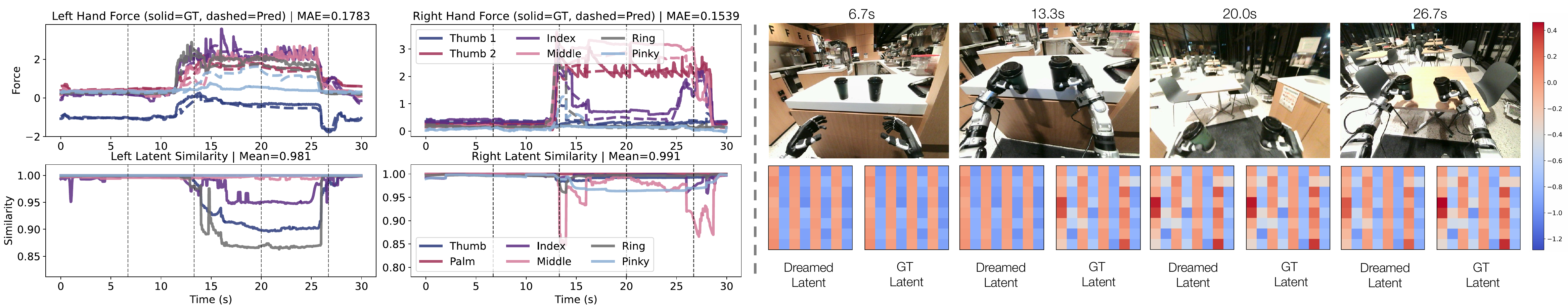}
        \caption{\textbf{Tea Serving.} Heatmaps with unnormalized latent values correspond to the \textbf{right middle} finger.}
        \label{fig:touch_dream_tea}
    \end{subfigure}
    
    \vspace{0.3em}
    
    \begin{subfigure}{1.0\linewidth}
        \includegraphics[width=\linewidth]{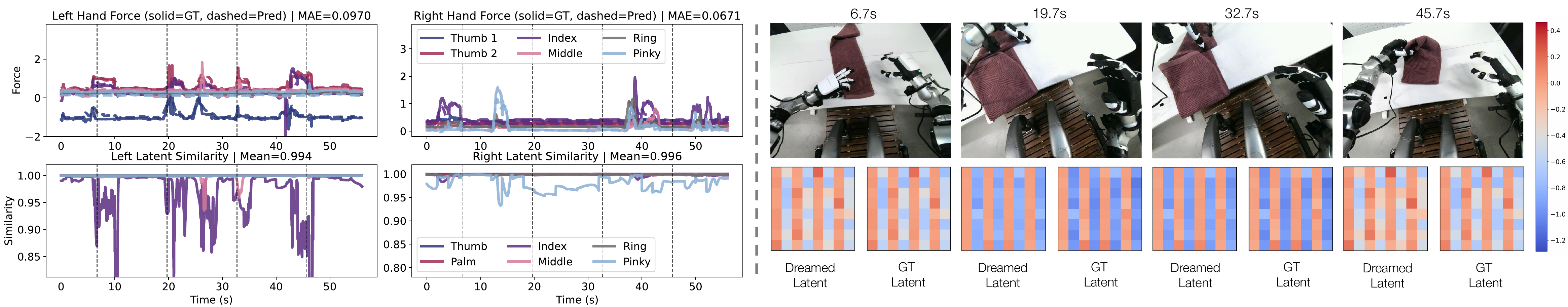}
        \caption{\textbf{Towel Folding.} Heatmaps with unnormalized latent values correspond to the \textbf{left index} finger.}
        \label{fig:touch_dream_towel}
    \end{subfigure}
    
    \caption{\textbf{Touch dreaming visualization.} We compare predicted (Pred) versus ground-truth (GT) future contact signals on representative rollouts for two tasks. For each task, the top left shows per-finger hand force trajectories and the mean absolute error (MAE) for the left and right hands, and the bottom left shows the corresponding normalized L2 similarity $s_{\mathrm{L2}}=1-\|\bm{p}-\bm{g}\|_2/(\|\bm{p}\|_2+\|\bm{g}\|_2)$ over time, where $\bm{p}$ and $\bm{g}$ denote predicted and ground-truth tactile latents, respectively. The vertical dashed lines in the left plots indicate the specific timestamps for the synchronized camera views and heatmaps of the dreamed versus ground-truth tactile latents shown on the right.}
    \label{fig:touch_dream}
\end{figure*}

\parab{Results and analysis.}
Fig.~\ref{fig:main_results} summarizes the main results. Across all five tasks, HTD consistently outperforms both decoder-only ACT baselines in both success rate and task score. Using the better-performing ACT baseline for each task and metric, HTD improves the task-averaged success rate and task score by 30.0 and 17.9 percentage points, corresponding to relative gains of about 90.9\% and 31.1\%, respectively. Importantly, simply adding force and tactile observations to ACT does not consistently improve performance: ACT (Visual + Proprio + Touch) outperforms ACT (Visual + Proprio) on only a subset of tasks and is not uniformly better in either metric. The largest gains of HTD appear on tasks that place stronger demands on contact-aware control or whole-body coordination. \textit{Insert-T} benefits from more accurate handling of tight-tolerance alignment and corrective contact. \textit{Towel Folding} highlights HTD's advantage on long-horizon deformable manipulation. \textit{Cat Litter Scooping} shows particularly large gains, reflecting the difficulty of combining tool use with squatting and constrained whole-body motion. In \textit{Tea Serving}, ACT often fails to rotate and move the body appropriately after successfully grasping both tea cups, whereas HTD is much more reliable. This likely reflects the importance of decoding low-dimensional but behavior-critical velocity commands with dedicated output tokens and independent action experts, rather than treating them as a small subset of a monolithic action vector. \textit{Book Organization} shows a smaller but still consistent gain, likely because it is more visually structured and has lower object-location variance. Overall, these results indicate that the full HTD framework is better suited than ACT baselines for versatile humanoid loco-manipulation.

\subsection{How do predictive touch dreaming and latent tactile supervision impact the effectiveness of contact-rich humanoid manipulation?}
\label{sec:exp_ablation}
To answer this question, we further isolate the contribution of touch observations and touch dreaming via ablations of HTD. 
Specifically, we evaluate four variants:
\textbf{w/o Touch and TD} removes both touch observations and the touch-dreaming objective. 
\textbf{w/o TD} keeps touch observations as input but removes the dreaming loss. 
\textbf{Dream Raw Tactile} predicts future tactile signals in the raw sensor space. 
\textbf{Dream Latent Tactile} (our full method) predicts future tactile signals in a learned latent space using supervision from an EMA teacher. 
All variants are trained with the same behavioral cloning objective for action chunk prediction, differing only in touch-related inputs and auxiliary objectives.


\parab{Results and analysis.}
Fig.~\ref{fig:ablation_results} reveals two main observations. First, \emph{touch as an input alone is not consistently beneficial}. Compared with w/o Touch and TD, w/o TD improves performance on Towel Folding but not Insert-T, slightly reduces average success rate, and yields only small task-score differences. Second, the benefits of predictive touch learning depend on the supervision space. Relative to w/o TD, both dreaming variants improve \textit{Insert-T} and average task score, whereas only Dream Latent Tactile also improves \textit{Towel Folding} and average success rate. Consequently, Dream Latent Tactile achieves the best overall performance, outperforming Dream Raw Tactile in both average metrics and yielding a 30\% relative gain in average success rate. Overall, these results show that touch inputs alone do not consistently improve performance and that latent-space supervision is more effective than raw tactile prediction.

\subsection{Does touch dreaming capture meaningful and robust contact-aware representations?}
To better understand what the touch-dreaming objective learns, we qualitatively examine whether the predicted future forces and tactile latents reflect meaningful contact dynamics during real-world rollouts. Fig.~\ref{fig:touch_dream} illustrates the dreamed touch representations across two policy rollout episodes, \textit{Tea Serving} and \textit{Towel Folding}. Across these representative rollouts, HTD predicts future hand force trajectories that effectively track both the timing and magnitude of contact events. Furthermore, tactile latent similarity (normalized L2 similarity) remains high during periods of sustained contact. On the right side of Fig.~\ref{fig:touch_dream}, we show the synchronized right-eye camera views together with unnormalized tactile-latent heatmaps for the selected fingers at the timesteps indicated by the dashed vertical lines in the plots on the left. We observe that the predicted heatmaps are close to the ground-truth heatmaps at most timesteps. While temporary drops in similarity occur during abrupt contact transitions and coincide with sudden force spikes, the similarity remains relatively high overall. The localized deviations are expected for two reasons. First, because the dreamed latent chunks are rolled out in an open-loop manner, the predictions naturally diverge slightly from the ground truth when unpredictable, discontinuous contact changes occur mid-chunk. Second, the ground-truth tactile latents typically vary with the noisy raw tactile readings, whereas the dreamed tactile latents are more stable and more aligned with semantically meaningful contact changes inferred jointly from vision and applied forces, rather than directly matching raw tactile readings that can be \textbf{noisy} or \textbf{spatially sparse}.

Comparing the two tasks further highlights the contact-awareness of the learned representations. Compared with the deformable-object interaction in \textit{Towel Folding}, \textit{Tea Serving} involves rigid objects and generally requires larger applied forces. During phases with light or sparse contact, such as \textit{Tea Serving} at 6.7\,s or the preliminary contacts in \textit{Towel Folding} at 19.7\,s and 32.7\,s, the latent patterns remain highly consistent across different fingers and tasks. Conversely, when richer contact occurs, the latents activate into distinct, high-intensity patterns, as seen in \textit{Tea Serving} at 13.3\,s, 20.0\,s, and 26.7\,s, and in \textit{Towel Folding} at 6.7\,s and 45.7\,s. Notably, when fingers experience comparable contact states, such as similar force magnitudes and localized contact regions, the resulting tactile latents exhibit visually analogous structural patterns. Meanwhile, the predicted tactile latents also show larger peak values under stronger contact and larger applied forces, for example when comparing the active contact patterns in \textit{Tea Serving} with those in \textit{Towel Folding}. This consistency across varying contact intensities suggests that the learned tactile latent space captures contact-relevant structure while being less affected by the spatial sparsity and local irregularity typical of raw sensor signals. Overall, these qualitative results reinforce the quantitative ablations by showing that the model learns a robust and contact-aware latent representation of physical interaction.

%% file: arxiv_contents/conclusions.tex

We present an integrated system for \emph{dexterous, contact-rich humanoid manipulation} combining a robust lower-body controller, a real-world VR-based humanoid demonstration pipeline, and a touch-aware policy learning framework. We propose \textbf{Humanoid Transformer with Touch Dreaming (HTD)}, a multimodal encoder--decoder Transformer that models touch as a core modality and augments behavioral cloning with future hand-joint-force and tactile-latent prediction. Supervision from an EMA target encoder enables stable, contact-aware latent learning without separate tactile pretraining.
Across five real-world tasks, \textit{Insert-T}, \textit{Book Organization}, \textit{Towel Folding}, \textit{Cat Litter Scooping}, and \textit{Tea Serving}, HTD consistently outperforms decoder-only ACT baselines, achieving a \textbf{90.9\%} relative improvement in average success rate over the stronger ACT variant for each task. Our ablations further show that latent tactile dreaming is more effective than raw tactile prediction, with Dream Latent Tactile yielding a \textbf{30\%} relative gain in success rate over Dream Raw Tactile. Together, these results suggest that combining robust whole-body execution, scalable humanoid data collection, and predictive touch-centered learning provides a practical path toward more reliable humanoid manipulation under frequent and complex contact changes.

%% file: arxiv_contents/appendix.tex
\subsection{Lower-Body Controller Details}
\label{sec:appendix_lbc}
We provide additional details on the command ranges and domain randomization parameters used in training the lower-body controller.

\paragraph{Command ranges and domain randomization}
Command signals are uniformly sampled from predefined ranges (Table~\ref{tab:cmd_ranges}), including the planar base velocity $\bm{v}_{xy}=[v_x, v_y]^\top$, base yaw rate $\omega_z$, torso orientation $\bm{rpy}=[\phi_{\text{torso}}, \theta_{\text{torso}}, \psi_{\text{torso}}]^\top$, and torso height $h$. In the reward equations below, a superscript ${}^\ast$ denotes the commanded target, while the corresponding unstarred quantity denotes the measured state. To improve sim-to-real transferability, we apply observation noise and randomize physics parameters (Table~\ref{tab:domain_rand}).

\begin{table}[h]
\centering
\footnotesize
\caption{Command ranges}
\label{tab:cmd_ranges}
\setlength{\tabcolsep}{4pt}
\renewcommand{\arraystretch}{0.9}
\begin{tabular}{lclc}
\toprule
\textbf{Command} & \textbf{Range} & \textbf{Command} & \textbf{Range} \\
\midrule
$v_x$ & $[-0.5,\;0.5]$ m/s & $\phi_{\text{torso}}$ (roll) & $[-0.7,\;0.7]$ rad \\
$v_y$ & $[-0.5,\;0.5]$ m/s & $\theta_{\text{torso}}$ (pitch) & $[-0.52,\;1.57]$ rad \\
$\omega_z$ & $[-1.57,\;1.57]$ rad/s & $\psi_{\text{torso}}$ (yaw) & $[-1.57,\;1.57]$ rad \\
$h$ & $[0.35,\;0.8]$ m & & \\
\bottomrule
\end{tabular}
\end{table}

\begin{table}[h]
\centering
\footnotesize
\caption{Domain randomizations}
\label{tab:domain_rand}
\setlength{\tabcolsep}{4pt}
\renewcommand{\arraystretch}{0.9}
\begin{tabular}{lclc}
\toprule
\textbf{Parameter} & \textbf{Range} & \textbf{Parameter} & \textbf{Range} \\
\midrule
Angular velocity & $\pm 0.2$ rad/s & Static friction & $[0.6,\;1.0]$ \\
Projected gravity & $\pm 0.05$ & Dynamic friction & $[0.4,\;0.8]$ \\
Joint position & $\pm 0.01$ rad & Restitution & $[0.0,\;0.005]$ \\
Joint velocity & $\pm 1.5$ rad/s & Base mass & $[-5.0,\;5.0]$ kg \\
\bottomrule
\end{tabular}
\end{table}

\paragraph{Achievable control ranges}
Our learned policy achieves the maximum stable controllable ranges summarized in Table~\ref{tab:wbc_range}, where the torso height is $[0.33,\,0.80]$\,m, torso roll is $[-0.38,\,0.35]$\,rad, torso pitch is $[-0.92,\,1.41]$\,rad, and torso yaw is $[-1.50,\,1.34]$\,rad. Compared with the command ranges used during training, the controller fully covers the trained height range and even slightly exceeds the lower bound, while covering most of the trained yaw range. For pitch, the stable range is broad but asymmetric: it extends beyond the trained range on the negative side, while being slightly smaller near the positive extreme. In contrast, the roll range is noticeably narrower than the training range, suggesting that lateral whole-body balance remains the most restrictive direction.

\begin{table}[h]
\centering
\caption{Maximum stable controllable ranges of our LBC policy in simulation.}
\label{tab:wbc_range}
\setlength{\tabcolsep}{6pt}
\renewcommand{\arraystretch}{1.0}
\begin{tabular}{lc}
\toprule
\textbf{Command} & \textbf{Stable range} \\
\midrule
Torso height $h$ & $[0.33,\,0.80]$ m \\
Torso roll $\phi_{\text{torso}}$ & $[-0.38,\,0.35]$ rad \\
Torso pitch $\theta_{\text{torso}}$ & $[-0.92,\,1.41]$ rad \\
Torso yaw $\psi_{\text{torso}}$ & $[-1.50,\,1.34]$ rad \\
\bottomrule
\end{tabular}
\end{table}

\subsection{Reward Details}

\providecommand{\ind}[1]{\ensuremath{\mathbf{1}\!\left[#1\right]}}
\providecommand{\pos}[1]{\ensuremath{\left[#1\right]_+}}
\providecommand{\clip}{\ensuremath{\mathrm{clip}}}

\newcommand{\rterm}[2]{%
\parbox[t]{\linewidth}{%
\raggedright\textbf{#1}\par
\vspace{1pt}%
\centering\scriptsize
\adjustbox{max width=0.98\linewidth}{$\textstyle #2$}%
}%
}

Table~\ref{tab:rewards_app} summarizes all reward terms used in training. The overall reward is composed of tracking rewards, regularization terms, contact- and gait-related terms, stability terms, and several auxiliary penalties. Below we briefly describe the role of each term.

\paragraph{Tracking rewards}
\begin{itemize}
    \item \textbf{Linear velocity reward.} Encourages the robot to track the commanded planar base velocity in the yaw-aligned horizontal frame, which is essential for stable omnidirectional locomotion.
    \item \textbf{Angular velocity reward.} Encourages accurate tracking of the commanded yaw angular velocity, allowing the robot to regulate turning behavior during locomotion.
    \item \textbf{Torso height reward.} Encourages the robot to maintain the desired torso height, which is important for both balance and workspace modulation during manipulation.
    \item \textbf{Torso roll reward.} Encourages tracking of the commanded torso roll in a horizontal frame aligned with the pelvis heading.
    \item \textbf{Torso pitch reward.} Encourages tracking of the commanded torso pitch in a horizontal frame aligned with the pelvis heading.
    \item \textbf{Torso yaw reward.} Encourages tracking of the commanded torso yaw relative to the pelvis, allowing the upper body to reorient independently for manipulation and coordination.
\end{itemize}

\paragraph{Regularization terms}
\begin{itemize}
    \item \textbf{Energy penalty.} Penalizes large instantaneous actuation effort to encourage energy-efficient motions and reduce unnecessary torque output.
    \item \textbf{Action rate penalty.} Penalizes rapid changes in consecutive actions, promoting smoother control signals and improving motion consistency.
    \item \textbf{Joint acceleration penalty.} Penalizes large joint accelerations to reduce abrupt motions and encourage physically plausible transitions.
    \item \textbf{Vertical velocity penalty.} Penalizes undesired vertical base motion, helping the robot maintain stable height regulation rather than bouncing during locomotion.
    \item \textbf{Roll/pitch rate penalty.} Penalizes excessive root angular velocity around the roll and pitch axes, expressed in the root body frame, reducing aggressive body oscillations and improving upper-body stability.
    \item \textbf{Joint torque limit penalties.} Penalize computed torques that exceed group-specific fractions of the actuator limits for the waist, ankle, and hip-pitch joints. We use normalized thresholds of $0.9$, $0.99$, and $0.99$, respectively.
\end{itemize}

\begin{table}[!t]
\centering
\footnotesize
\caption{Reward terms}
\label{tab:rewards_app}
\setlength{\tabcolsep}{1pt}
\renewcommand{\arraystretch}{1.05}

\begin{tabular}{@{}
>{\raggedright\arraybackslash}p{0.38\columnwidth}
>{\centering\arraybackslash}p{0.10\columnwidth}
>{\raggedright\arraybackslash}p{0.38\columnwidth}
>{\centering\arraybackslash}p{0.10\columnwidth}
@{}}
\toprule
\textbf{Term} & \textbf{Weight} & \textbf{Term} & \textbf{Weight} \\
\midrule
\multicolumn{4}{c}{Tracking rewards} \\
\midrule

\rterm{Linear velocity}{
r_{\text{vel}} := \exp\!\left(-\frac{\|\mathbf{v}_{xy}-\mathbf{v}_{xy}^\ast\|_2^2}{\sigma_{v}^2}\right)
}
& 1.0
&
\rterm{Torso roll}{
\begin{aligned}
r_{\text{roll}} &:= \exp\!\left(-\frac{(\phi_{p_y}-\phi^\ast)^2}{\sigma_{r}^2}\right),\\
\widetilde{\mathbf q}_{t} &:= \mathbf q_{\mathrm{yaw}}(\mathbf q_p)^{-1}\otimes\mathbf q_t,\\
\phi_{p_y} &:= \mathrm{wrap}_{[-\pi,\pi]}\!\left(\mathrm{roll}(\widetilde{\mathbf q}_{t})\right)
\end{aligned}
}
& 1.0
\\

\rterm{Angular velocity}{
r_{\text{ang}} := \exp\!\left(-\frac{(\omega_z-\omega_z^\ast)^2}{\sigma_{\omega}^2}\right)
}
& 1.0
&
\rterm{Torso pitch}{
\begin{aligned}
r_{\text{pitch}} &:= \exp\!\left(-\frac{(\theta_{p_y}-\theta^\ast)^2}{\sigma_{p}^2}\right),\\
\widetilde{\mathbf q}_{t} &:= \mathbf q_{\mathrm{yaw}}(\mathbf q_p)^{-1}\otimes\mathbf q_t,\\
\theta_{p_y} &:= \mathrm{wrap}_{[-\pi,\pi]}\!\left(\mathrm{pitch}(\widetilde{\mathbf q}_{t})\right)
\end{aligned}
}
& 1.5
\\

\rterm{Torso height}{
r_h := \exp\!\left(-\frac{(h-h^\ast)^2}{\sigma_h^2}\right)
}
& 1.0
&
\rterm{Torso yaw}{
\begin{aligned}
r_{\text{yaw}} &:= \exp\!\left(-\frac{(\Delta \psi-\psi^\ast)^2}{\sigma_{y}^2}\right),\\
\Delta \psi &:= \psi_{\text{torso}}-\psi_{\text{pelvis}}
\end{aligned}
}
& 1.0
\\

\midrule
\multicolumn{4}{c}{Regularization} \\
\midrule

\rterm{Energy}{
r_E := \left\|\, |\boldsymbol{\tau}\odot \dot{\mathbf{q}}| \,\right\|_2
}
& $-0.001$
&
\rterm{Action rate}{
r_{\Delta a} := \|\mathbf{a}_t-\mathbf{a}_{t-1}\|_2^2
}
& $-0.01$
\\

\rterm{Joint acceleration}{
r_{\ddot{q}} := \sum_{j\in \mathcal{J}_{\text{lower}}} \ddot{q}_j^2
}
& $-2.5\times 10^{-7}$
&
\rterm{Vertical velocity}{
r_{v_z} := v_z^2
}
& $-1.0$
\\

\rterm{Roll/pitch rate}{
r_{\omega_{xy}} := \omega_{b,x}^2+\omega_{b,y}^2
}
& $-0.10$
&
\rterm{Joint torque-limit penalties}{
\begin{aligned}
r_{\tau\text{-lim}}(\mathcal{J}_g)
:=\sum_{j\in\mathcal{J}_g}
\max\!\left(
\frac{|\tau_j^{\mathrm{comp}}|}{\tau_j^{\max}}-\lambda_g,\,0
\right)
\end{aligned}
}
& \shortstack{$-100$\\each}
\\

\midrule
\multicolumn{4}{c}{Contact \& Gait} \\
\midrule

\rterm{Undesired contacts}{
r_{\text{uc}} := \sum_{b\in \mathcal{B}_{\text{nonfoot}}}
\mathbb{I}\!\left(\max_{\tau\in[t-\Delta,t]}\|\mathbf{F}_b(\tau)\|_2 > F_{\text{thr}}\right)
}
& $-1.0$
&
\rterm{Feet slide}{
r_{\text{slide}} := \sum_{f\in\{L,R\}} \|\mathbf{v}^{xy}_{f}\|_2 \cdot \mathbb{I}(f\ \text{in contact})
}
& $-0.25$
\\

\rterm{Flying}{
r_{\text{fly}} := \mathbb{I}\!\left(\text{no foot contact}\right)
}
& $-1.0$
&
\rterm{Feet force}{
\begin{aligned}
F_z^{\mathrm{agg}} &:= \sqrt{F_{z,L}^{2}+F_{z,R}^{2}},\\
r_{F} &:= \clip\!\left(\max(F_z^{\mathrm{agg}}-500,\,0),\,0,\,400\right)
\end{aligned}
}
& $-0.003$
\\

\rterm{Feet air-time}{
\begin{aligned}
t_f^{\mathrm{mode}} &:=
\begin{cases}
t_f^{\mathrm{con}}, & f\text{ in contact},\\
t_f^{\mathrm{air}}, & \text{otherwise},
\end{cases}\\
r_{\text{air}}
:=\;& \min\!\left(\min_{f\in\{L,R\}}t_f^{\mathrm{mode}},\,0.4\right)
\mathbb{I}(\text{single-stance})\\
&\cdot \mathbb{I}\!\left(\|\mathbf{v}_{xy}^\ast\|_2 + |\omega_z^\ast| > 0.1\right)
\end{aligned}
}
& $0.15$
&
\rterm{Feet stumble}{
\begin{aligned}
r_{\text{stumble}}
&:= \mathbb{I}\!\left(
\exists f\in\{L,R\}:\ 
\|\mathbf{F}^{xy}_{f}\|_2 > 5\,|F_{z,f}|
\right)
\end{aligned}
}
& $-2.0$
\\

\rterm{Standing contact}{
\begin{aligned}
r_{\text{stand}}
:=\;&
\tanh\!\left(\frac{\min_f t_f^{\mathrm{con}}}{0.5}\right)
\mathbb{I}\!\left(\min_f t_f^{\mathrm{con}}>0.5\right)\\
&\cdot
\mathbb{I}\!\left(\|\mathbf{v}_{xy}^{\ast}\|_2+|\omega_z^{\ast}|\leq 0.1\right)
\end{aligned}
}
& $0.5$
& &
\\

\midrule
\multicolumn{4}{c}{Stability} \\
\midrule

\rterm{Joint limits}{
\begin{aligned}
r_{\text{lim}}
:= \sum_{j}
\Big(
\max(q_j-q^{\max}_j,\,0)
+\max(q^{\min}_j-q_j,\,0)
\Big)
\end{aligned}
}
& $-2.0$
&
\rterm{Flat orientation}{
r_{\text{flat}} := \| \mathbf{g}_{xy} \|_2^2
}
& $-2.0$
\\

\rterm{Feet distance}{
r_{\text{near}}
:= \max\!\left(d_{\text{thr}}-\|\mathbf{p}_L-\mathbf{p}_R\|_2,\,0\right)
}
& $-2.0$
&
\rterm{Center of gravity}{
r_{\text{CoG}}
:=
\exp\!\left(
-\frac{
\left\|
\mathbf{p}_{\text{CoG}}^{xy}
-\frac{1}{2}\left(\mathbf{p}_{L}^{xy}+\mathbf{p}_{R}^{xy}\right)
\right\|_2^2
}{0.2^2}
\right)
}
& $0.5$
\\

\rterm{Knee stance width}{
r_{\text{knee}}
:=
\exp\!\left(
-\frac{
\left(
\|\mathbf{p}_{k,L}^{xy}-\mathbf{p}_{k,R}^{xy}\|_2-0.27
\right)^2
}{0.1^2}
\right)
}
& $0.5$
& &
\\

\midrule
\multicolumn{4}{c}{Other} \\
\midrule

\rterm{Joint deviation}{
r_{\text{dev}}(\mathcal{J})
:= \sum_{j\in\mathcal{J}} |q_j-q^{\text{default}}_j|
}
& $-0.5$ to $-0.02$
&
\rterm{Termination}{
r_{\text{term}} := \mathbb{I}(\text{terminated})
}
& $-200$
\\
\bottomrule
\end{tabular}
\end{table}

\paragraph{Contact and gait terms}
\begin{itemize}
    \item \textbf{Undesired contacts penalty.} Penalizes collisions involving non-foot body parts, encouraging the robot to avoid falling, dragging limbs, or contacting the environment with inappropriate links.
    \item \textbf{Feet slide penalty.} Penalizes foot motion while in contact with the ground, encouraging stable footholds and reducing slip.
    \item \textbf{Flying penalty.} Penalizes states in which no foot is in contact with the ground, discouraging unstable airborne phases during normal locomotion.
    \item \textbf{Feet force penalty.} Penalizes the combined magnitude of the two feet's vertical ground reaction forces above $500$\,N, with the excess clipped at $400$\,N.
    \item \textbf{Feet air-time reward.} Rewards sustained single-stance phases under non-trivial motion commands using the shorter of the stance foot's contact time and the swing foot's air time.
    \item \textbf{Feet stumble penalty.} Penalizes abnormal contact patterns in which tangential foot force becomes excessively large relative to vertical support force, which often indicates stumbling or unstable foot-ground interaction.
    \item \textbf{Standing contact reward.} Rewards sustained bilateral foot contact during near-zero planar-velocity and yaw-rate commands, promoting stable stationary postures.
\end{itemize}

\paragraph{Stability terms}
\begin{itemize}
    \item \textbf{Joint limits penalty.} Penalizes violations of soft joint position limits, preventing the policy from relying on unrealistic or unsafe joint configurations.
    \item \textbf{Flat orientation penalty.} Penalizes non-flat base orientation, encouraging upright and balanced locomotion behavior.
    \item \textbf{Feet distance penalty.} Penalizes configurations in which the two feet become too close to each other, which helps maintain a reasonable support polygon and improves balance robustness.
    \item \textbf{Center-of-gravity reward.} Rewards alignment of the mass-weighted whole-body center of gravity with the midpoint between the two ankles, improving balance.
    \item \textbf{Knee stance-width reward.} Encourages the horizontal distance between the knees to remain near $0.27$\,m, reducing inward knee collapse during height and posture changes.
\end{itemize}

\paragraph{Other terms}
\begin{itemize}
    \item \textbf{Joint deviation penalty.} Penalizes deviation from default joint configurations for selected joint groups, with weights $-0.5$ for hip-yaw/hip-roll joints, $-0.2$ for waist joints, and $-0.02$ for the remaining leg joints.
    \item \textbf{Termination penalty.} Applies a large penalty when the episode terminates, strongly discouraging catastrophic failure such as falling or unrecoverable instability.
\end{itemize}